\documentclass[lettersize,journal]{IEEEtran}
\usepackage{amsmath,amsfonts}
\usepackage{algorithmic}
\usepackage{algorithm}
\usepackage{array}
\usepackage{textcomp}
\usepackage{stfloats}
\usepackage{url}
\usepackage{verbatim}
\usepackage{graphicx}
\usepackage{cite}
\usepackage{booktabs}
\usepackage{microtype}
\usepackage{csquotes}
\usepackage{authblk}
\usepackage{multirow}
\usepackage{makecell}
\usepackage{enumitem}
\usepackage[labelformat=simple]{subcaption}

\usepackage[dvipsnames]{xcolor}

\begin{document}

\title{Self-Supervised Generative-Contrastive Learning of Multi-Modal Euclidean Input for 3D Shape Latent Representations: A Dynamic Switching Approach}

\author[1]{Chengzhi Wu}
\author[2]{Julius Pfrommer}
\author[1]{Mingyuan Zhou}
\author[1,2]{Jürgen Beyerer}
\affil[1]{Institute for Anthropomatics and Robotics, Karlsruhe Institute of Technology, Germany}
\affil[2]{Fraunhofer Institute of Optronics, System Technologies and Image Exploitation IOSB, Germany\vspace{.5\baselineskip}}
\affil[ ]{\tt \small chengzhi.wu@kit.edu \qquad julius.pfrommer@iosb.fraunhofer.de}
\affil[ ]{\tt \small mingyuan.zhou@student.kit.edu \qquad juergen.beyerer@iosb.fraunhofer.de}


\markboth{Journal of \LaTeX\ Class Files,~Vol.~xx, No.~xx, November~2023}%
{Shell \MakeLowercase{\textit{et al.}}: A Sample Article Using IEEEtran.cls for IEEE Journals}


\maketitle

\begin{abstract}
We propose a combined generative and contrastive neural architecture for learning latent representations of 3D volumetric shapes. The architecture uses two encoder branches for voxel grids and multi-view images from the same underlying shape. The main idea is to combine a contrastive loss between the resulting latent representations with an additional reconstruction loss. That helps to avoid collapsing the latent representations as a trivial solution for minimizing the contrastive loss. A novel dynamic switching approach is used to cross-train two encoders with a shared decoder. The switching approach also enables the stop gradient operation on a random branch. Further classification experiments show that the latent representations learned with our self-supervised method integrate more useful information from the additional input data implicitly, thus leading to better reconstruction and classification performance.
\end{abstract}


\section{Introduction}
\label{sec:introduction}
\IEEEPARstart{3}{D} shapes can be represented in a range of different formats. On the Euclidean side, they may be represented as RGB-D images, multi-view images or volumetric data. On the Non-Euclidean side, they may be represented as point clouds or meshes. For computer vision tasks like classification, segmentation, or even generative tasks like shape reconstruction, the target 3D shape is usually converted into a latent representation first. 
Before the rise of deep learning \cite{Goodfellow2015DeepL}, popular latent representations (or, 3D shape descriptors) were Laplacian spectral eigenvectors \cite{SorkineHornung2005LaplacianMP}, or heat kernel signatures \cite{Sun2009ACA}. With neural networks, the latent representation is usually the result of an encoder that reduces the 3D shape to a vector representation with fixed dimensionality.

\begin{figure}[t]
    \centering
    \begin{subfigure}[b]{\linewidth}
        \centering
        \includegraphics[width=0.95\linewidth,trim=2 2 2 2,clip]{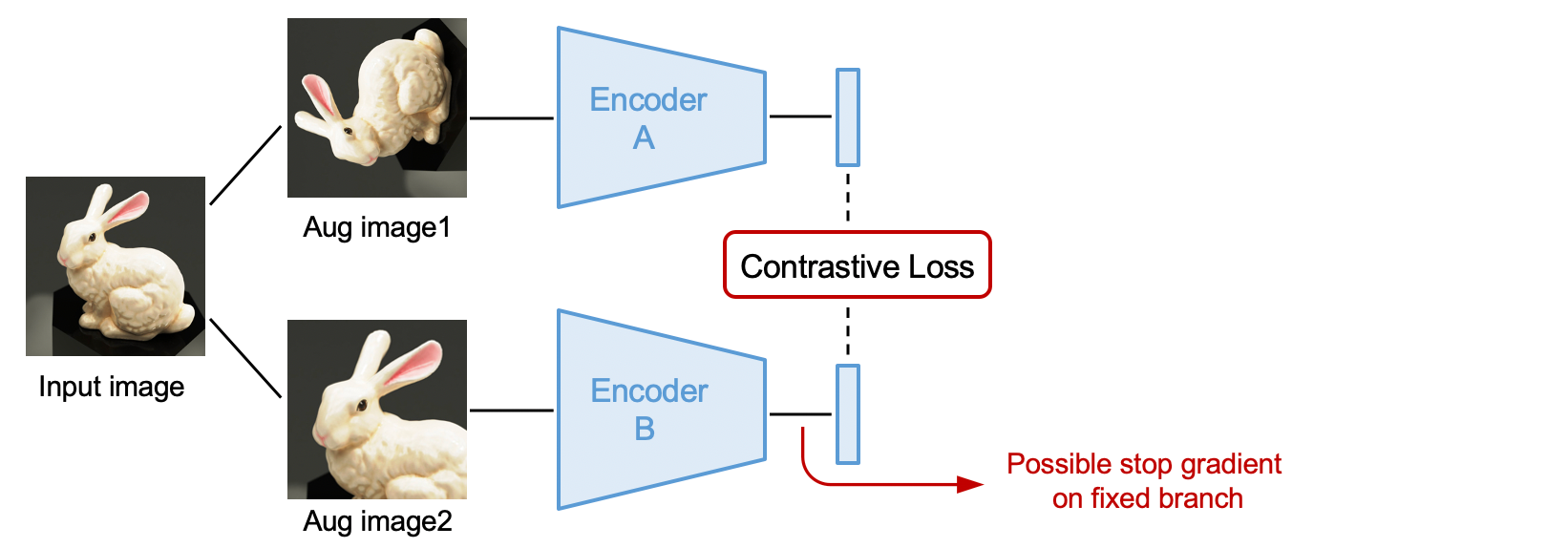}
        \caption{General pipeline for contrastive learning from augmented data.}
        \label{fig:ContrasL}
    \end{subfigure}
    \begin{subfigure}[b]{\linewidth}
        \centering
        \includegraphics[width=0.95\linewidth,trim=2 2 2 2,clip]{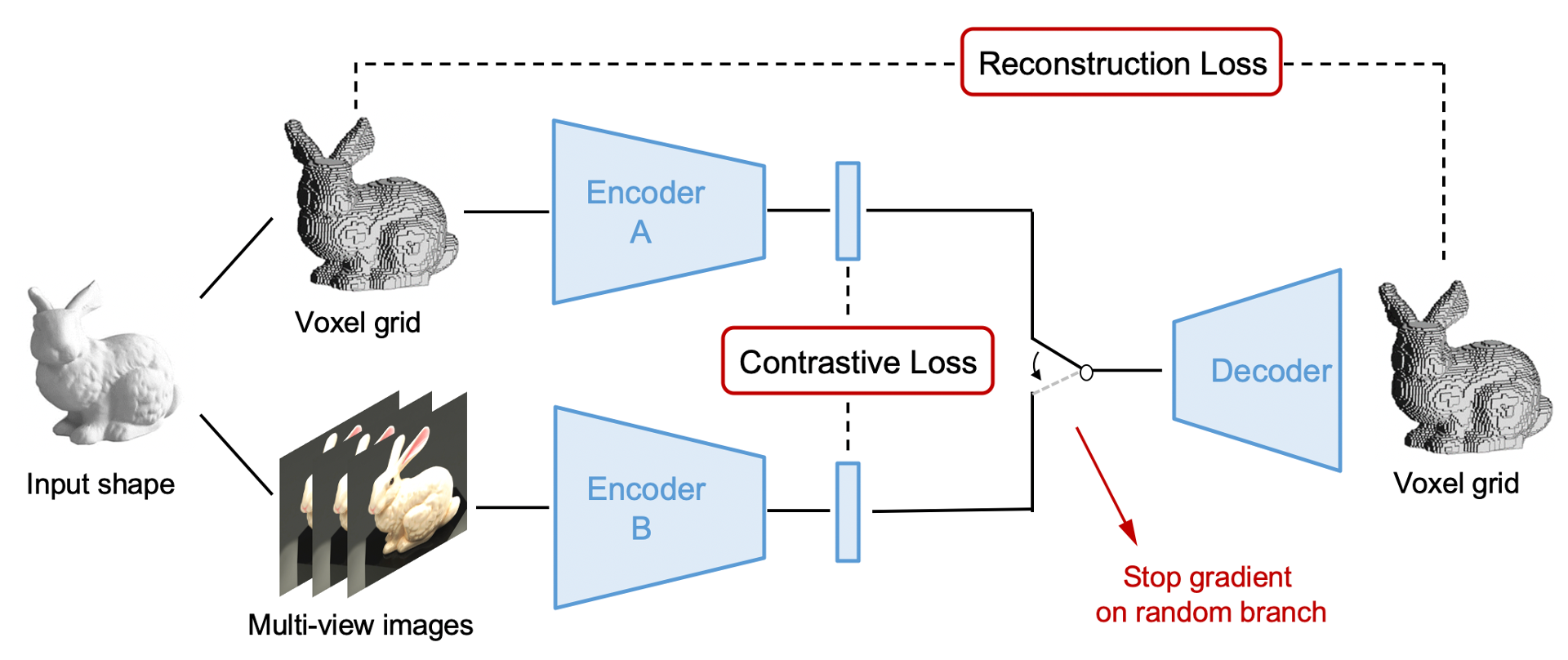}
        \caption{Pipeline for the proposed generative-contrastive learning from multi-modal input.}
        \label{fig:GenContrasL}
    \end{subfigure}
    \caption{An illustration of (a) a general pipeline of contrastive learning methods and (b) our proposed generative contrastive learning pipeline for 3D shapes.}
    \label{fig:ConL&GenConL}
\end{figure}

When multi-modal input data is available, the question arises of how to use them jointly. For 3D learning tasks, take 3D Euclidean data as an example, most state-of-the-art methods in computer vision that deal with both image and voxel grid input data either concatenate individual latent representations for supervised tasks \cite{Hegde2016FusionNet3O}, or use only one of them on the input and loss side separately \cite{Choy20163DR2N2AU, Wu2016LearningAP, Tulsiani2018MultiviewCA, Yan2016PerspectiveTN}, or use them jointly but with pre-training and finetuning \cite{Girdhar2016LearningAP}. We are interested in seeking a better self-supervised way for learning better latent representations for 3D volumetric shapes, with additional input from other modalities.

Apart from pretext tasks-based methods\cite{Gidaris2018UnsupervisedRL, Goyal2019ScalingAB}, the other two main self-supervised learning ways are generative-based methods \cite{Brock2016GenerativeAD, Choy20163DR2N2AU} and contrastive-based methods \cite{Chen2020ASF, Grill2020BootstrapYO}. For 3D volumetric shapes, it is easy to implement a generative model. But it is still an open question of how to do it in a contrastive way, let alone the combination of these two. In a recent review paper of self-supervised learning \cite{Liu2020SelfsupervisedLG}, the authors argue that the only way of doing generative-contrastive learning is to train an encoder-decoder to generate synthetic samples and a discriminator to distinguish them from real samples. We disagree with this argument. In their definition, the discriminator is the contrastive part thus the model only focuses on negative pairs. We think it is also possible to use or only use positive pairs, e.g. in our case, using multi-modalities from the same input shape for two branches.

Figure \ref{fig:ConL&GenConL} shows the main idea of our proposed generative-contrastive learning pipeline. Compared to the existing contrastive learning methods, our method shares some similarities with them while some significant differences also exist. Similarities are: (i) we both use a two-branches scheme to encode two inputs that originate from the same "raw data"; (ii) after getting encoded latent representations, we both compute a contrastive loss in the latent space; (iii) they use positive pairs for training (optionally with additional negative pairs), we also use positive pairs. Differences are: (i) they use different augmented data from the "raw data", while we use different modalities from the "raw data"; (ii) thus our network architectures of encoder A and B are not identical, while theirs are identical mostly; (iii) we add a decoder part and a reconstruction loss; (iv) they possibly have stop gradient on one fixed branch, while we do stop gradient on random branch with a switching approach.

The main contributions of this paper are as follows:
\begin{itemize}[itemsep=0pt,topsep=1pt,left=0pt]
\item We propose a novel generative-contrastive learning pipeline for 3D volumetric shapes, which makes the joint training of encoders for multi-modal input data possible.
\item With the switching approach doing the work of stopping gradient on random branch, model collapse is avoided. End-to-end training is also possible without the requirements of special pre-training.
\item Using the voxel encoder as a self-supervised pre-trained feature extractor, we outperform 3D-GAN on the ModelNet40 classification task with much shorter latent vector representations (128, compared to ca. 2.5 million dimensions in 3D-GAN).
\item The voxel encoder pre-trained on one single category still performs surprisingly well as a feature extractor on the full dataset with other categories during the testing.
\end{itemize}

\section{Related Work}
\label{sec:relatedWork}
\textbf{Contrastive learning: }
The work of contrastive learning was pioneered by Yann LeCun's group for face verification \cite{Chopra2005LearningAS}. This topic has been getting more and more popular recently since people find self-supervised learning is important for feature extraction and we now have really mature deep learning techniques. SimCLR \cite{Chen2020ASF} proposes to use two identical encoders for two branches, both positive pairs and negative pairs are used. MoCo \cite{Chen2020ImprovedBW} stops the gradient for the second branch, while using a momentum-based method to update the parameters of its encoder. SwAV \cite{Caron2020UnsupervisedLO} proposes to use a memory bank to get negative pairs out of the batch, the contrastive loss in their case is computed after clustering. For methods that only use positive samples, BYOL \cite{Grill2020BootstrapYO} keeps the idea of momentum updating from MoCo, but adds an additional block in the first branch and only uses positive pairs. SimSiam \cite{Chen2020ExploringSS} reports an observation of competitive results may still be achieved when modifying BYOL by making two encoders identical. A review of the most relevant methods and their comparisons are given in \cite{Ericsson2020HowWD}. For 3D data, contrastive learning-based frameworks have been proposed mainly for the point cloud data representation, e.g. PointContrast \cite{Xie2020PointContrastUP} and Contrastive Scene Contexts \cite{Hou2020ExploringD3}. More recently, several contrastive learning frameworks have been proposed for multi-modality input. Most of them focus on text-to-image learning \cite{Zhang2021CrossModalCL, Li2020UNIMOTU, Ye2022CrossmodalCL, Zolfaghari2021CrossCLRCC}. For 3D shapes, the closest work to ours is CrossPoint \cite{Afham2022CrossPointSC}, which uses images and point clouds as input for better point cloud latent representation learning. However, it only uses the contrastive loss, no decoder or reconstruction loss is used. A more detailed comparison is given in subsection \ref{sec:crosspoint}.

\textbf{Learning on 3D shapes with Euclidean data: }
For supervised tasks, VoxNet\cite{Maturana2015VoxNetA3} is the pioneer in using 3D convolutional network to learn features from volumetric data for recognition. Its subsequent work of multi-level 3D CNN \cite{Ghadai2019MultiLevel3C} learns multi-scale spatial features by considering multiple resolutions of the voxel input. Qi et al. \cite{Qi2016VolumetricAM} propose to use multi-resolution filtering in 3D for multi-view CNNs, as well as using subvolume supervision for auxiliary training. FusionNet \cite{Hegde2016FusionNet3O} fuses three networks together: two VoxNets\cite{Maturana2015VoxNetA3} and one MVCNN\cite{Su2015MultiviewCN}. The three networks fuse at the score layers where a linear combination of scores is taken before the classification prediction. A more recent work of Simple3D-Former \cite{Wang2022CanWS} uses all image, voxel, and point cloud data as the input and co-trains the framework with multi-tasks in a supervised manner.

For self-supervised tasks, ShapeNet \cite{Wu20153DSA} uses a reverse VoxNet to reconstruct 3D volumetric shapes from latent representations that are learned from depth maps. The T-L network \cite{Girdhar2016LearningAP} combines a 3D autoencoder with an image regressor to encode a unified vector representation given a 2D image. Autoencoders have also been widely use for 3D shape retrieval in other papers \cite{Xie2015DeepshapeDL, Zhu2014DeepLR}. Its variant, VAE, has been used in a similar way for 3D shape learning \cite{Brock2016GenerativeAD}. View information from images has also been widely investigated for 3D shape reconstruction. Choy et al. \cite{Choy20163DR2N2AU} proposed a framework named 3D-R2N2 to reconstruct 3D shapes from multi-view images by leveraging the power of recurrent neural networks \cite{Hochreiter1997LongSM}. \cite{Rezende2016UnsupervisedLO} also uses a recurrent-based approach, but taking depth images as input. Some other methods use view information as auxiliary constraints \cite{Tulsiani2018MultiviewCA, Yan2016PerspectiveTN, Gwak2017WeaklyS3}. Method uses GAN for 3D volumetric shape generation has been proposed in \cite{Wu2016LearningAP}. Some other latest works \cite{Muralikrishnan2019ShapeUA, Klokov2019ProbabilisticRN} have also used multi-modal input data for joint end-to-end training, but they did not use a switching approach for dynamic training.

We are aware that there are lots of other works applying deep learning-based methods on other 3D Non-Euclidean data formats, e.g. point clouds \cite{Pang2022MaskedAF, Xie2020PointContrastUP, Hou2020ExploringD3, Afham2022CrossPointSC, Qian2022Pix4PointIP, Liu2021LearningF2}. However, it should be acknowledged that these methods have been primarily designed and optimized for point cloud data, and as such, their encoders are mostly not plug-and-play modules. Several critical intricacies, such as the sampling of point cloud patches, and the processing of the perspective-variant point cloud input, cannot be directly transferred to volumetric data. (An exception is CrossPoint \cite{Afham2022CrossPointSC}, whose encoder is a plug-and-play module and we can easily replace it with a voxel encoder for volumetric data training.) Overall, they are out of the scope of this work, but we plan to include other 3D non-Euclidean data representations, e.g. point cloud, in our future work so that we could perform a fairer comparison with them directly.

\begin{figure}[t]
    \centering
    \includegraphics[width=\linewidth]{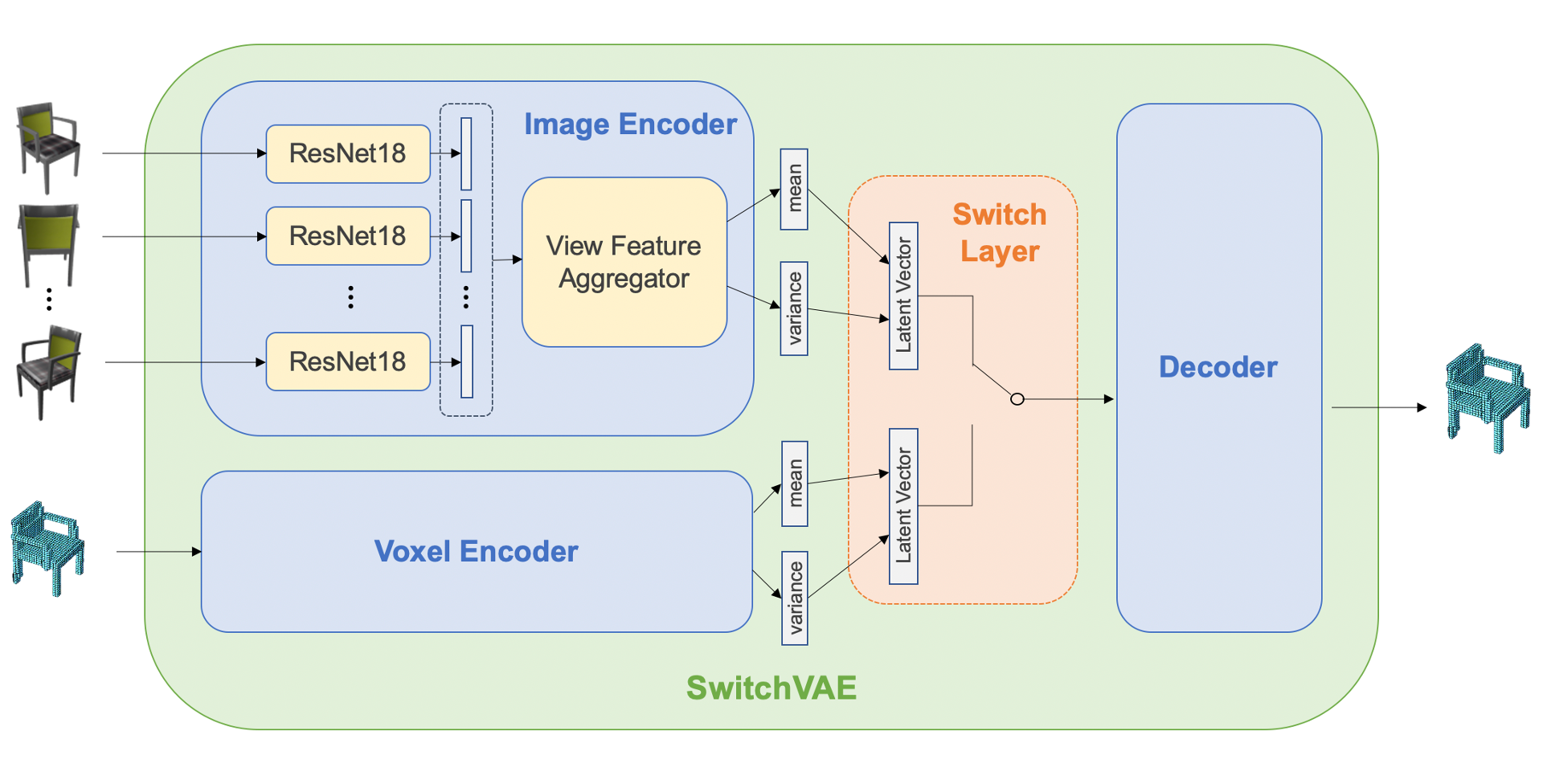}
    \caption{The SwitchVAE architecture based on our proposed generative-contrastive learning pipeline.}
    \label{fig:switchVAE}
\end{figure}

\section{Methodology}
\label{sec:methodology}

\subsection{Generative-Contrastive Learning}
Figure \ref{fig:GenContrasL} shows the main idea of our proposed generative-contrastive learning pipeline. Similar to most existing contrastive learning methods, we use an architecture with two encoder branches to compute a contrastive loss between the latent representations from each branch. The inputs to the two branches are the voxel grid and the multi-view images of an identical 3D shape. A generative decoder part is added to compute the reconstruction loss. The decoder is shared by two encoders and the two encoder branches are co-trained with the help of a switching approach. 

For contrastive learning methods, mode collapse is a big issue. Possible ways of dealing with it are adding additional blocks for encoder A, or stopping gradient for encoder B and updating its parameters in a momentum way slowly along with the updated parameters in encoder A. In our case, encoders A and B are already different network architectures thus the momentum method can not be applied, but we still managed to avoid model collapse successfully during the experiments. We attribute this success to two things: the reconstruction loss, and the switching approach. The reconstruction loss has a strong supervision over the representational capacity of latent representations, while the switching approach does the work of stopping the gradient on random branch.

To further improve the latent representations, Variational Autoencoders (VAE) \cite{Kingma2019AnIT} are used instead of vanilla autoencoders. In a VAE, each input is mapped to a multivariate normal distribution around a point in the latent space, which makes a continuous latent space. A continuous latent space makes the smooth transition of 3D shapes possible with latent representations. The learned features are usually more smooth and meaningful.

\subsection{Switch Encoding}
When dealing with multi-modal inputs, most state-of-the-art methods just encode them separately into latent representations and then perform concatenation. Unlike them, we propose to use a switching approach in the latent space to jointly train both encoders with a shared decoder. During the training, the switch is actuated for every training epoch with a preset probability to randomly select the encoded output from one encoder as the latent representation. This operation of switching between encoders continues during the whole end-to-end training.

The decoder is tasked to reconstruct the voxel representation of the 3D shape. Since the switched encoders are trained concurrently for the same decoder, they are forced to produce \enquote{mutually compatible} latent representations. The different input modalities result in different features that naturally emerge for the respective latent representation. For example, the voxel encoder tends to generate the latent feature of the full shape, while the image encoder can only generate the latent feature based on specific views, which may lose information due to occlusions. By cross-training with switched encoding, useful features for the latent representation can be translated from one encoder to the other via the shared decoder. This results in improved latent representations also for the individual encoder when just one input modality is used after the training.

\begin{figure*}[t]
    \centering
    \begin{subfigure}[b]{0.36\linewidth}
        \centering
        \begin{tabular}{ccc}
        Input/GT & Voxel VAE & SwitchVAE \\ \hline
        \end{tabular}
        \includegraphics[width=\linewidth]{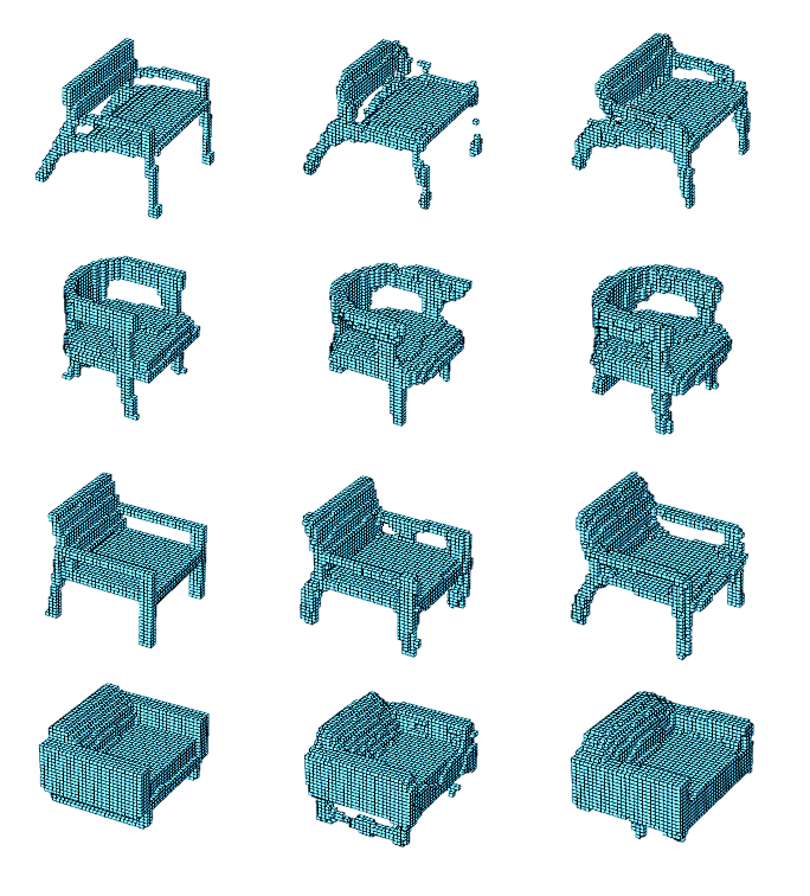}
        \caption{Reconstruction tests with voxel input.}
        \label{fig:recon_1}
    \end{subfigure}
    \quad\quad
    \begin{subfigure}[b]{0.53\linewidth}
        \centering
        \begin{tabular}{p{2.6cm}<{\centering} p{1.4cm}<{\centering} cc}
        Input (2/8 views) & GT & Image VAE & SwitchVAE \\ \hline
        \end{tabular}
        \includegraphics[width=\linewidth]{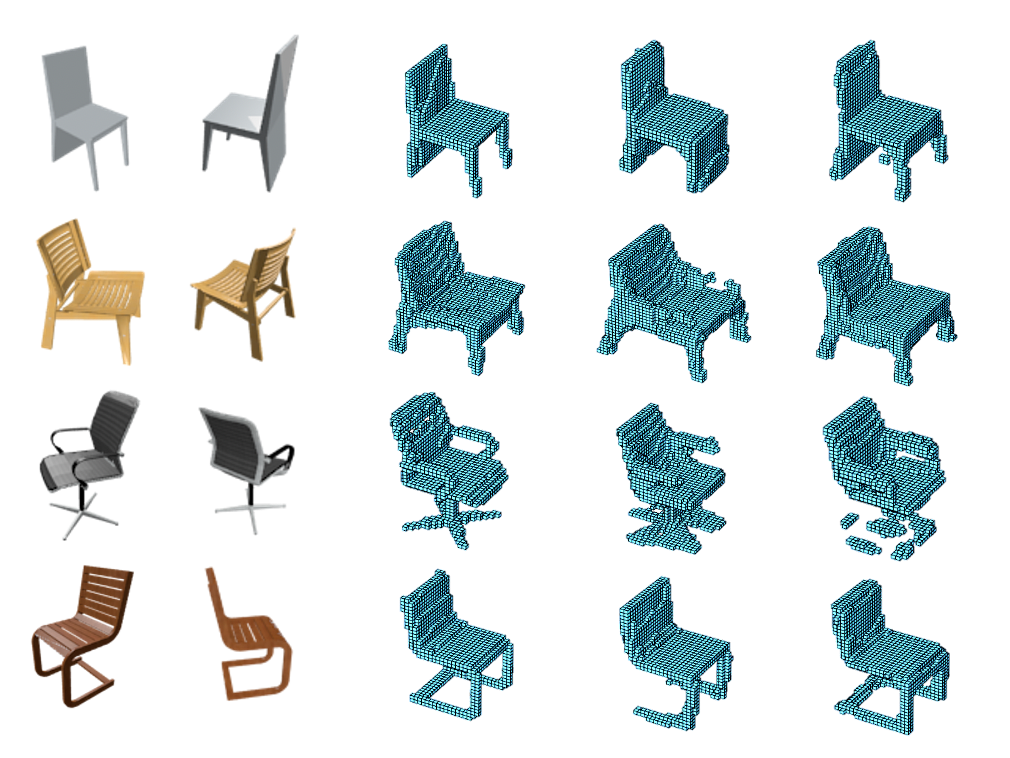}
        \caption{Reconstruction tests with image input.}
        \label{fig:recon_2}
    \end{subfigure}
    \caption{Some reconstruction results from different models with only voxel or multi-view images as the test input.}
    \label{fig:recon}
\end{figure*}

\begin{table*}[t]
\centering
\begin{tabular}{cc p{2cm}<{\centering} p{2cm}<{\centering} p{2cm}<{\centering}}
\toprule
\multirow{2}{*}{\textbf{Test Input}} & \multirow{2}{*}{\textbf{Training Model}} & \multicolumn{3}{c}{\textbf{Reconstruction Metrics}} \\ \cmidrule{3-5}
 &  & IoU & Precision & Accuracy\\ \midrule
\multirow{3}{*}{Image} & Image VAE & 58.52\% & 68.21\%  & 93.21\% \\
 & SwitchVAE ($\lambda_{\mathrm{contras}}=0$) & 56.82\% & 67.72\% & 93.00\% \\
 & SwitchVAE ($\lambda_{\mathrm{contras}}=1$) & \textbf{58.75\%} & \textbf{68.50\%} & \textbf{93.32\%} \\ \midrule
\multirow{3}{*}{Voxel} & Voxel VAE & 78.86\% & 82.80\% & 97.01\% \\
 & SwitchVAE ($\lambda_{\mathrm{contras}}=0$) & 77.27\% & 80.04\% & 96.67\%\\
 & SwitchVAE ($\lambda_{\mathrm{contras}}=1$) & \textbf{79.93\%} & \textbf{84.68\%} & \textbf{97.22\%} \\ \bottomrule
\end{tabular}
\caption{Reconstruction performance on the test set. Training and testing with the chair category from 3D-R2N2 dataset.}
\label{table:1}
\end{table*}

\subsection{Loss Functions}
The SwitchVAE loss function consists of three parts: a reconstruction loss $L_{\mathrm{recon}}$, a KL divergence $L_{\mathrm{KL}}$ between latent representations and the normal prior distribution, and a contrastive loss $L_{\mathrm{contras}}$ between the latent representations from the different input formats. The overall network is parameterized by $\boldsymbol \theta = (\boldsymbol \theta_{\mathrm{vox}}, \boldsymbol \theta_{\mathrm{img}}, \boldsymbol \theta_d)^{\top}$ for the voxel and image encoder and the voxel decoder respectively. The training samples are denoted $\boldsymbol x^\alpha$ for the input $\alpha\in\{\mathrm{img},\mathrm{vox}\}$. The switch value for $\alpha$ is randomly selected prior to every training epoch. The latent representations resulting from the VAE encoders are $(\boldsymbol \mu^\alpha, \boldsymbol \sigma^\alpha) = e^\alpha(\boldsymbol x^\alpha)$. The latent representation is sampled for the current training epoch as $\boldsymbol z^{\alpha} \sim \mathcal N(\boldsymbol \mu^\alpha, \boldsymbol \sigma^\alpha)$. The decoder part is shared by both input modalities to reconstruct the voxel representation $\hat{\boldsymbol x}^\alpha = d(\boldsymbol z^\alpha)$. Formally, the overall loss function decomposes into three terms
\begin{equation}
\begin{aligned}
    & L_{\boldsymbol\theta}(\alpha, \boldsymbol x^{\mathrm{img}}, \boldsymbol x^{\mathrm{vox}}) = {} \\ & \qquad L_{\mathrm{recon}}(\boldsymbol x^{\mathrm{vox}}, \hat{\boldsymbol x}^\alpha) + \lambda_{\mathrm{KL}} L_{\mathrm{KL}}(\boldsymbol \mu ^\alpha, \boldsymbol \sigma^\alpha) + {}\\
    & \qquad \lambda_{\mathrm{contras}} L_{\mathrm{contras}}(\boldsymbol z^{\mathrm{img}}, \boldsymbol z^{\mathrm{vox}})
    \end{aligned}
    \label{equ:loss}
\end{equation}
with weights $\lambda_{\mathrm{KL}}$ and $\lambda_{\mathrm{contras}}$.

A modified Binary Cross Entropy (BCE) against the voxel ground truth is used for the reconstruction loss. To improve the training, modification has been made by the introduction of a hyper-parameter $\gamma$ that weights the relative importance of false positives against false negatives. The reconstructed voxels are indexed by $k$ with value $\hat x^\alpha_k \in [0,1]$.
\begin{equation}
\begin{aligned}
    L_{\mathrm{recon}}(\boldsymbol x^{\mathrm{vox}}, \hat{\boldsymbol x}^\alpha) &{}= \sum_k \Big[-\gamma \cdot x^{\mathrm{vox}}_k \cdot \log(\hat x^\alpha_k)-{}\\[-.25\baselineskip]
    & (1-\gamma)(1-x^{\mathrm{vox}}_k)\log(1-\hat x^\alpha_k) \Big]
    \end{aligned}
\end{equation}
We set the hyperparameter $\gamma=0.8$ during training for all of the experiments conducted in Section \ref{sec:experiments}.

In the training of VAE, the Kullback-Leibler (KL) divergence is used between the actual distribution of latent vectors and the $\mathcal N(\boldsymbol 0, I)$ Gaussian distribution. Note that the latent representation has $n$ dimensions.
\begin{equation}
    L_{\mathrm{KL}}(\boldsymbol \mu, \boldsymbol \sigma) = 
    -\frac{1}{2}\sum_{i=1}^n (1 + \log(\sigma_{i}^2)-\mu_{i}^2-\sigma_{i}^2) 
\end{equation}

In order to further force a close distance between the latent representations learned from image and volumetric data with the SwitchVAE model, a contrastive loss between the encoders is proposed and used in the latent space during the training phase. The contrastive loss is defined as the Euclidean distance between the latent vectors from images and volumetric data.
\begin{equation}
    L_{\mathrm{contras}}(\boldsymbol z^{\mathrm{img}}, \boldsymbol z^{\mathrm{vox}}) = \| \boldsymbol z^{\mathrm{img}} - \boldsymbol z^{\mathrm{vox}} \|_2^2
\end{equation}
Although in most other contrastive learning methods some different contrastive losses has been used, e.g. InfoNCE loss in SimCLR \cite{Chen2020ASF}, we find that with latent representations normalized, our method can already yield satisfying results with a simple $L_2$ Norm loss as the contrastive loss.

\section{Experiments}
\label{sec:experiments}
We use the 3D-R2N2 and ModelNet 10/40 datasets for our experiments. The 3D-R2N2 dataset \cite{Choy20163DR2N2AU} is a subset with 13 categories from the ShapeNet dataset \cite{Chang2015ShapeNetAI}. It provides good quality rendered multi-view images alongside a class label and $32\times 32\times 32$ voxel representations. We divide the 3D-R2N2 dataset into a training set of 29,599 samples and a test set of 7406 samples. The ModelNet dataset \cite{Wu20153DSA} comes in two variations with either 10 or 40 classes of shapes. The ModelNet10 dataset contains 3991/908 training/test samples. ModelNet40 contains 9843/2468 training/test samples.

For the SwitchVAE models, we use both a voxel and a multi-view image encoder. The decoder always reconstructs the voxel representation. During training for voxel test input, the switch layer randomly selects either the voxel encoder with a probability of 80\%, or the multi-view image encoder with a probability of 20\%.

Concerning the other training parameters, we use a latent dimension of 128 for all experiments. The network parameters are trained by minimizing the loss function from Equation \ref{equ:loss} using the SGD optimizer with a momentum of 0.9 and Nesterov accelerated gradients \cite{Ruder2016AnOO}. The learning rate is $2\times 10^{-4}$ with a decay of 0.96 per 10 epochs after the first 50 epochs. The batch size is 32 for all experiments. Training with multi-view image input uses 8 views for every sample as it has been reported in \cite{Choy20163DR2N2AU} and its subsequent works \cite{Xie2019Pix2VoxC3, Xie2020Pix2VoxMC} that the improvement from additional views is negligible after the first 6-10 views.

\begin{table*}[t]
\centering
\begin{tabular}{ccc p{2cm}<{\centering} p{2cm}<{\centering}}
\toprule
\multirow{2}{*}{\textbf{Training Dataset}} & \multirow{2}{*}{\textbf{Test Input}} & \multirow{2}{*}{\textbf{Training Method}} & \multicolumn{2}{c}{\textbf{Classification Accuracy}} \\ \cmidrule{4-5} 
 &  &  & ModelNet40 & ModelNet10 \\ \midrule
\multirow{4}{*}{Chair Category} & \multirow{2}{*}{Multi-view images} & Image VAE & 75.28\% & 81.36\% \\
 &  & SwitchVAE & \textbf{77.07\%} & \textbf{84.26\%} \\ \cmidrule{2-5} 
 & \multirow{2}{*}{Voxel data} & Voxel VAE & 80.19\% & 86.38\% \\
 &  & SwitchVAE & \textbf{80.60\%} & \textbf{87.05\%} \\ \midrule
\multirow{4}{*}{ModelNet40} & \multirow{2}{*}{Multi-view images} &    Image VAE & \textbf{85.06\%} & 88.62\% \\ 
 &  & SwitchVAE & 83.87\% & \textbf{89.96\%} \\ \cmidrule{2-5} 
 & \multirow{2}{*}{Voxel data} & Voxel VAE & 83.12\% & 87.95\% \\
 &  & SwitchVAE & \textbf{84.01\%} & \textbf{90.07\%} \\ \bottomrule
\end{tabular}
\caption{Classification accuracy on the ModelNet40/ModelNet10 classification tasks with models trained with Image VAE, Voxel VAE, and SwitchVAE on the chair category or on the full ModelNet40 dataset.}
\label{table:2}
\end{table*}

\begin{table*}[t]
\centering
\begin{tabular}{ccccc}
\toprule
\multirow{2}{*}{\textbf{Supervision}} & \multirow{2}{*}{\textbf{Method}} & \multirow{2}{*}{\textbf{Data Modality}} & \multicolumn{2}{c}{\textbf{Classification Accuracy}} \\ \cmidrule{4-5} 
& & & ModelNet40 & ModelNet10 \\ \midrule
\multirow{6}{*}{Supervised} & 3D ShapNets \cite{Wu20153DSA} & Voxels & 77.30\% & 85.30\% \\
 & VoxNet \cite{Maturana2015VoxNetA3} & Voxels & 83.00\% & 92.00\% \\
 & MVCNN \cite{Su2015MultiviewCN} & Images & 90.10\% & - \\
 & FusionNets \cite{Hegde2016FusionNet3O} & Images, Voxels & 90.80\% & 93.11\% \\
 & 3D2SeqViews \cite{Han20193D2SeqViewsAS} & Images & 93.40\% & 94.71\% \\
 & VRN Ensemble \cite{Brock2016GenerativeAD} & Voxels & 95.54\% & 97.14\% \\ \midrule
\multirow{7}{*}{Self-supervised} 
 & LFD \cite{Chen2003OnVS} & Images & 75.50\% & 79.90\% \\
 & T-L Network \cite{Girdhar2016LearningAP} & Images, Voxels & 74.40\% & - \\
 & VConv-DAE \cite{Sharma2016VConvDAEDV} & Voxels & 75.50\% & 80.50\% \\
 & 3D GAN \cite{Wu2016LearningAP} & Voxels & 83.30\% & \textbf{91.00\%} \\
 & CrossVoxel (modified from CrossPoint \cite{Afham2022CrossPointSC}) & Images, Voxels & 78.82\% & 86.34\% \\
 & SwitchVAE (trained on only chair category) & Images, Voxels & 80.60\% & 87.05\% \\
 & SwitchVAE (trained on ModelNet40 dataset) & Images, Voxels & \textbf{84.01\%} & 90.07\% \\ 
\bottomrule
\end{tabular}
\caption{Classification accuracy on the ModelNet40/ModelNet10 dataset with different methods. Results from methods that only used images and/or voxels are listed. Note that our latent representations are only of 128 dimensions.}
\label{table:3}
\end{table*}

\subsection{Detailed Network Configuration}
Figure \ref{fig:switchVAE} shows based on our proposed generative-contrastive learning pipeline, how switched encoding is implemented for a VAE with multi-view images and voxel grids input. The encoder blocks of our SwitchVAE build on the idea of volumetric convolutional networks \cite{Wu20153DSA} for the voxel input, and 3D recurrent reconstruction neural networks \cite{Choy20163DR2N2AU} for the multi-view images input. More detailed network configurations are given as follows.

The image encoder of SwitchVAE learns the latent vector from multi-view images, and it is composed of a view feature embedding module and a view feature aggregator module. The view feature embedding module is a ResNet18 \cite{He2016DeepRL} whose weights are shared across all the views. The part with pre-trained weights maps a single view $137\times 137\times 3$ RGB image into $5\times 5\times 512$ feature maps. We then flatten these feature maps and add a fully connected layer, which outputs a 1024-dimensional feature for a single view image. For a 3D shape, 8 views of images are fed into the shared weights view feature embedding module while training, which outputs the $8\times 1024$ view features. 

For the view feature aggregator module, we first tried max pooling as MVCNN \cite{Su2015MultiviewCN}, but it did not yield satisfying results. Same for average pooling. To better aggregate the multi-view image features, we finally use the Gated Recurrent Unit (GRU) \cite{Cho2014LearningPR}. The view feature aggregator outputs a 1024-dimensional feature after aggregating features from all views. Then it is further fed into the last fully connected layers to generate the mean and the variance of the latent vector. By using the reparametrization trick introduced in \cite{Kingma2014AutoEncodingVB}, the image encoder finally outputs a 128-dimensional sampled latent vector.

The voxel encoder is a 3D volumetric convolutional neural network. The encoder has 4 convolutional layers and two fully connected layers. All convolutional layers use kernels of size $3\times 3\times 3$, their strides are $\{1, 2, 1, 2\}$ and channel numbers are $\{8, 16, 32, 64\}$ respectively. All layers use the exponential linear unit (eLu) as the activation function except for the last fully connected layer. This layer maps a shape of $32\times 32\times 32$ voxels to a 343-dimensional feature. The 343-dimensional feature is further fed into the last fully connected layers to generate the mean and the variance of the latent vector to finally produce a 128-dimensional sampled latent vector.

The decoder of SwitchVAE mirrors the voxel encoder, except that the last layer uses a sigmoid activation function. The decoder maps a 128-dimensional latent vector, which was randomly sampled in the encoder, to a $32\times 32\times 32$ volumetric reconstruction. It represents the predicted voxel occupancy possibility of each voxel in the cube.

\subsection{3D Shape Reconstruction}
We use Intersection-of-Union (IoU), precision, recall, and accuracy (referred to as average precision in \cite{Girdhar2016LearningAP}) as the quantitative metrics for the reconstruction of 3D shapes. The threshold at which a voxel is considered as filled is 50\%. Similar to the last part, we show the results from only voxel input training, only image input training, and both input training with our SwitchVAE.

Table \ref{table:1} shows that the reconstruction performance of SwitchVAE is similar or slightly better to that of image/voxel VAEs, and it focuses more on making every predicted occupied voxel correct (higher precision score). This characteristic may be more clearly observed in some reconstruction results. Table \ref{table:1} also shows that the contrastive loss term is the key in our method. Figure \ref{fig:recon} shows some qualitative reconstruction results from our SwitchVAE model that trained on the chair category. Comparisons with the results from the networks that only use one input format for training are also presented. From the figure we can observe that SwitchVAE takes more attention on not occupying the original negative voxels. This is quite obvious from the third row of Figure \ref{fig:recon_2}. Both Image VAE and SwitchVAE are not certain about the leg number of the office chair is 4 or 5. The Image VAE decides to merge them all together, while SwitchVAE decides to only guess and occupy some voxels with small sub-clusters in that area.

\begin{figure}[t]
    \centering
    \begin{subfigure}[b]{0.48\linewidth}
        \centering
        \includegraphics[width=\linewidth]{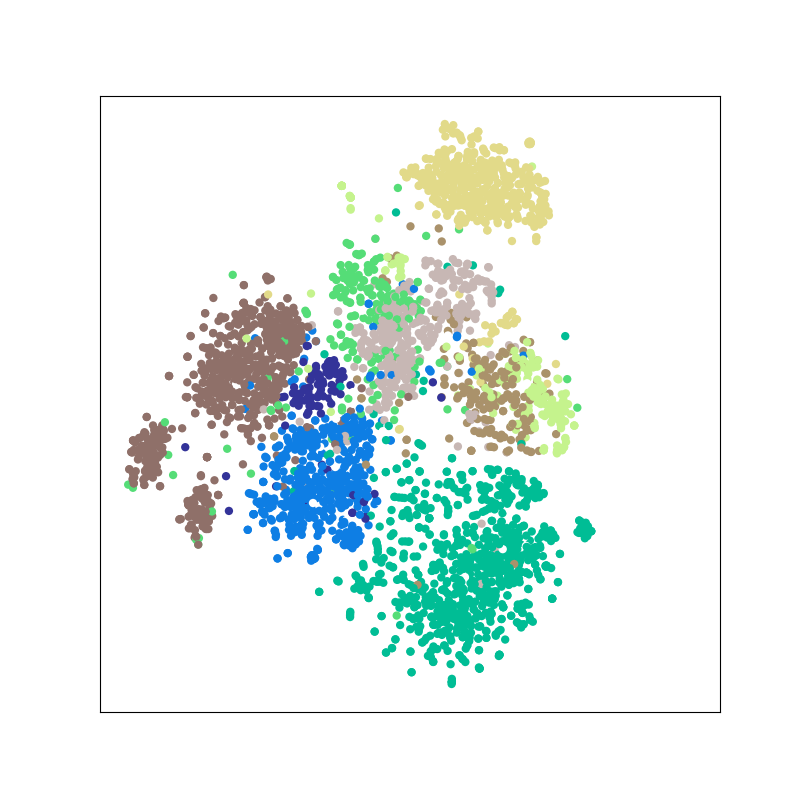}
        \caption{\centering \ \ Voxel VAE, \newline trained on ModelNet10}
        \label{fig:tsne_1}
    \end{subfigure}
    \begin{subfigure}[b]{0.48\linewidth}
        \centering
        \includegraphics[width=\linewidth]{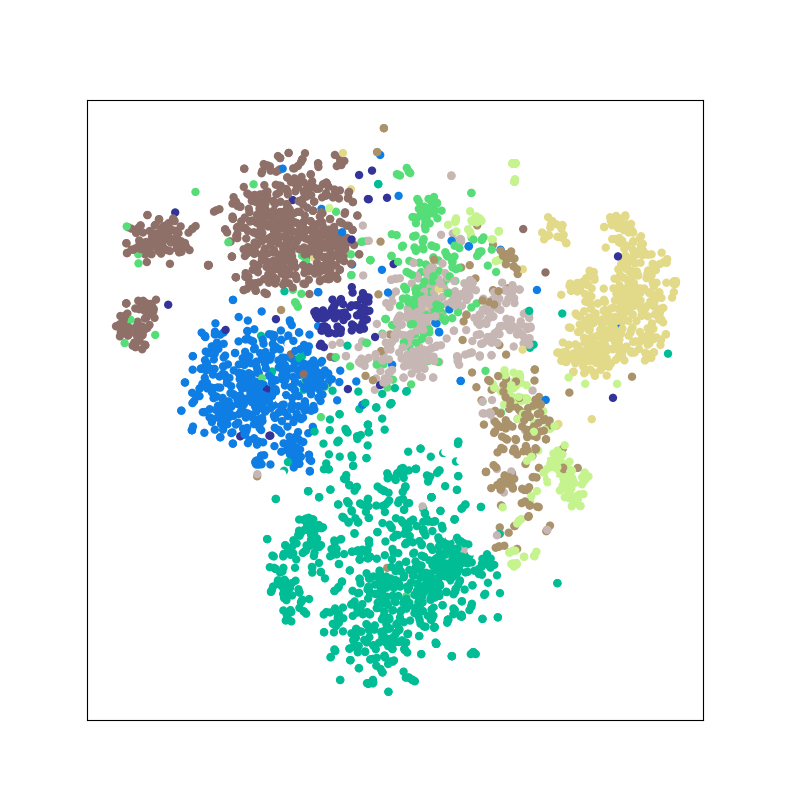}
        \caption{\centering SwitchVAE ($\lambda_{\mathrm{contras}}=0$), \newline trained on ModelNet10}
        \label{fig:tsne_2}
    \end{subfigure}
    \begin{subfigure}[b]{0.48\linewidth}
        \centering
        \includegraphics[width=\linewidth]{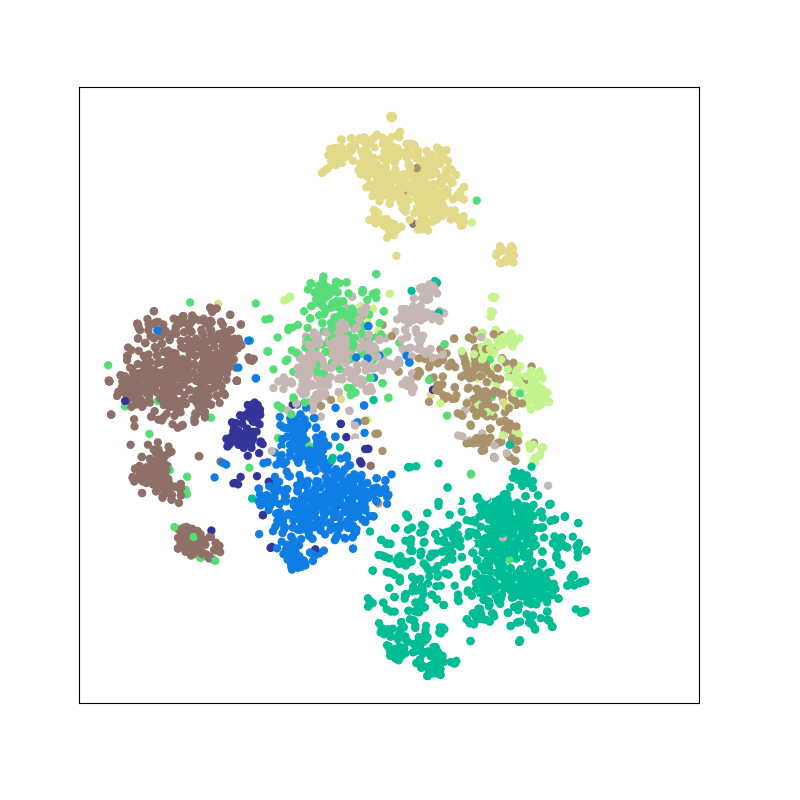}
        \caption{\centering SwitchVAE ($\lambda_{\mathrm{contras}}=1$), \newline trained on ModelNet10}
        \label{fig:tsne_3}
    \end{subfigure}
    \begin{subfigure}[b]{0.48\linewidth}
        \centering
        \includegraphics[width=\linewidth]{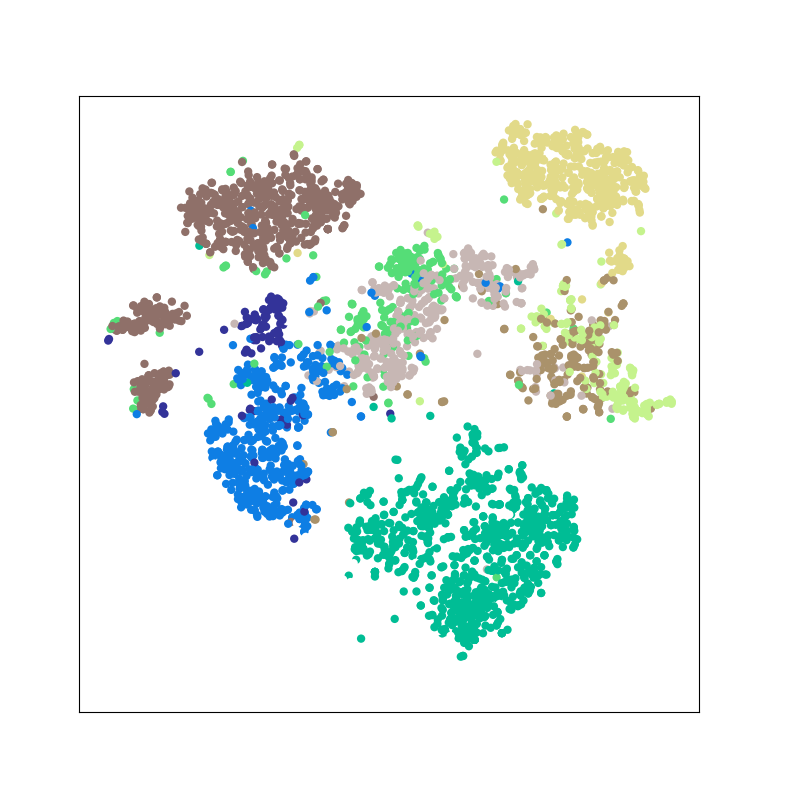}
        \caption{\centering SwitchVAE ($\lambda_{\mathrm{contras}}=1$), \newline trained on 3D-R2N2 dataset}
        \label{fig:tsne_4}
    \end{subfigure}
    \caption{t-SNE plots of the latent representations for ModelNet10 shapes (10 categories) with (a) a vanilla voxel VAE model trained on ModelNet10 dataset. (b) a SwitchVAE model without contrastive loss trained on ModelNet10 dataset. (c) a SwitchVAE model with contrastive loss trained on ModelNet10 dataset. (d) a SwitchVAE model with contrastive loss trained on the full 3D-R2N2 dataset. Each color represents one category. All latent representations used for the plots use voxel data from the testing set as test input.}
    \label{fig:tsne}
\end{figure}

\subsection{3D Shape Classification} 
\label{sec:cls}
For the classification task, the networks are first trained to perform reconstruction of the ground-truth voxel representations. Then the encoder part of a trained network is used to produce latent representations of 128 dimensions as input for classification. An SVM with RBF kernel and hyper-parameter $\gamma=1/128$ is trained on the latent representations to perform classification. The same samples were used to train the networks for latent representations and the SVM. The evaluation of the SVM is performed with samples that were neither used to train the networks nor the SVM.

Table~\ref{table:2} shows the impact of switched training on the ModelNet 10/40 classification tasks. To make it more clear, let's take the voxel data tests as an example. During the training phase, a voxel VAE only trains on the voxel train set data, while a SwitchVAE trains on both the voxel train set data and the correspondent multi-view images train set data. During the testing phase, only the identical voxel test set data is given to the trained voxel VAE model and the trained SwitchVAE model. Latent representations of those 3D shapes (from the voxel test set) obtained from the SwitchVAE model always outperform that from the vanilla voxel VAE in the ModelNet classification tasks. Note that no multi-view image data of the test set is needed for SwitchVAE during the testing. From Table~\ref{table:2}, we can clearly observe that under the condition of the same training dataset and test input, the results from SwitchVAE are better than the results from the image VAE or the voxel VAE in most cases. Note that they are even trained with a same number of epochs. This means by using the data of other formats in the training phase, during the testing phase, the classification performance has been improved compared to the models that only use a single format for the training.  

Table~\ref{table:3} lists the classification result in comparison to other network architectures. Note that there are not many papers on image-voxel multi-modal methods for 3D shapes. Hence, we have included some other methods that used images or voxels solely as input for additional comparison. Compared to most other unsupervised learning method, we achieve better classification performance. Compared to 3D-GAN, our method outperforms it on the ModelNet40 classification task and achieves competitive performance on the ModelNet10 classification task. However, our method uses a much smaller latent vector of only 128 dimensions. 3D-GANs use all feature maps in the last three convolution layers, which makes the presentation for each 3D shape a 2.5 million dimensional vector as input for the classification.

\begin{figure}[t]
    \centering
    \includegraphics[width=0.95\linewidth,trim=4 4 4 4,clip]{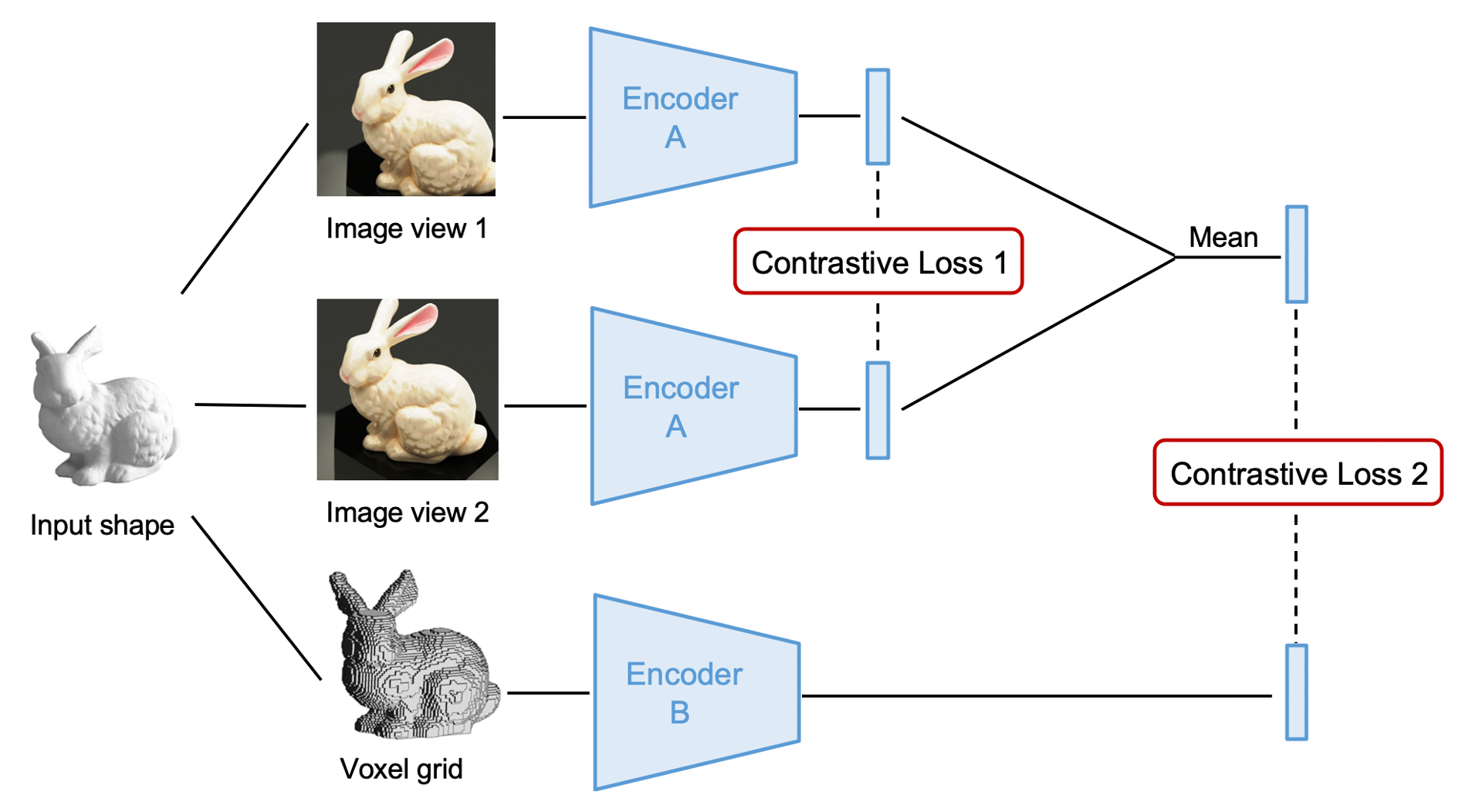}
    \caption{The framework of CrossVoxel, which is modified from CrossPoint \cite{Afham2022CrossPointSC} for comparison expermiments.}
    \label{fig:crosspoint}
\end{figure}

\textbf{t-SNE Visualization:} 
In order to visualize the learned latent representations, we use t-SNE \cite{Maaten2008VisualizingDU} to map the latent representations to a 2D plane. Figure \ref{fig:tsne} gives the visualization results. we use ModelNet10 for most t-SNE visualization experiments. Comparing Figure \ref{fig:tsne_1} and Figure \ref{fig:tsne_2}, we can observe that the switching approach contributes to the inter-category classification while making the intra-category clustering a bit fuzzy (all categories are a bit far from each other, while each category itself is a bit less clustered). Comparing Figure \ref{fig:tsne_2} and Figure \ref{fig:tsne_3}, we can observe that adding the contrastive loss term to SwitchVAE helps the intra-category clustering, making the performance of the whole classification task much better. Comparing Figure \ref{fig:tsne_3} and Figure \ref{fig:tsne_4}, we can observe that with a larger training dataset, even better feature clustering results may be achieved. The increased gaps between different categories can be clearly observed.

\begin{table}[t]
\centering
\resizebox{1\linewidth}{!}{
\begin{tabular}{ccccc}
\toprule
\multirow{2}{*}{\textbf{Method}} & \multirow{2}{*}{\begin{tabular}[c]{@{}c@{}}\textbf{Pre-train}\\ \textbf{(unsupervised)}\end{tabular}} & \multirow{2}{*}{\begin{tabular}[c]{@{}c@{}}\textbf{Fine-tune} \\ \textbf{(supervised)}\end{tabular}} & \multicolumn{2}{c}{\textbf{Classification Accuracy}} \\ \cmidrule{4-5} 
& & & ModelNet40 & ModelNet10 \\ \midrule
Fully supervised & $\times$ & $\checkmark$ & 89.74\% & 92.32\% \\
SwithVAE & $\checkmark$ & $\times$ & 84.01\% & 90.07\% \\
SwithVAE (fine-tuned) & $\checkmark$ & $\checkmark$ & \textbf{91.25\%} & \textbf{93.40\%} \\
\bottomrule
\end{tabular}}
\caption{Experimental results of further fine-tuning the pre-trained voxel encoder with a simple classification head in a supervised manner.}
\label{table:4}
\end{table}

\begin{table}[t]
\centering
\begin{tabular}{ccc}
\toprule
\multirow{2}{*}{\begin{tabular}[c]{@{}c@{}}\textbf{Switch Probability} \\ \textbf{(image : voxel)}\end{tabular}} & \multicolumn{2}{c}{\textbf{Classification Accuracy}} \\ \cmidrule{2-3} 
& ModelNet40 & ModelNet10 \\ \midrule
0 : 10 & 83.12\% & 87.95\% \\
1 : 9 & 83.47\% & 89.22\% \\
2 : 8 & \textbf{84.01\%} & \textbf{90.07\%} \\
3 : 7 & 83.92\% & 89.84\% \\
4 : 6 & 83.21\% & 88.19\% \\
5 : 5 & 82.63\% & 86.70\% \\
\bottomrule
\end{tabular}
\caption{Ablation study on switch probability. Using voxel data as the test data.}
\label{table:5}
\end{table}

\textbf{Fine-tuned Classification:} 
We additionally report the result of pre-train the encoder in a self-supervised manner with our SwitchVAE, then fine-tune it supervised with a simple classification head. The numerical results are reported in Table \ref{table:4}. For a better comparison, the result of the encoder directly trained supervised with the classification head is also reported. From the table, we can see that the voxel encoder can further improve its performance when pre-trained with our method before the training of supervised learning.

\begin{figure}[t]
    \centering
    \begin{subfigure}[b]{\linewidth}
        \centering
        \includegraphics[width=\linewidth,trim=4 4 4 4,clip]{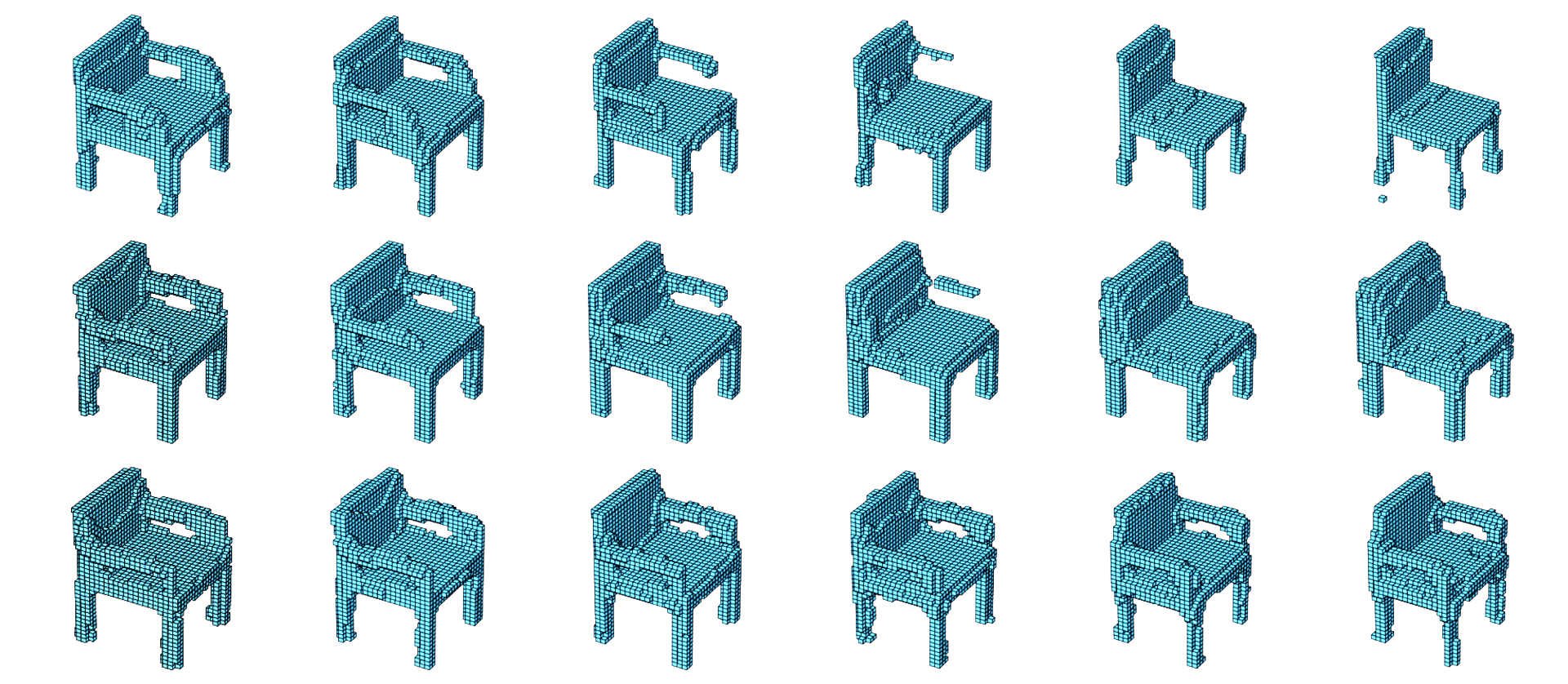}
        \caption{Top Row: Trained on SwitchVAE, the \enquote{chair arm} feature and the \enquote{size} feature are entangled. Middle and Bottom Row: Trained on Switch-BetaTCVAE with $\beta=5$, the \enquote{chair arm} feature and the \enquote{size} feature are more disentangled, changing one feature does not impact the other one too much.}
        \label{fig:btcvae1}
    \end{subfigure}
    \begin{subfigure}[b]{\linewidth}
        \centering
        \vspace{0.2cm}
        \includegraphics[width=\linewidth,trim=4 4 4 4,clip]{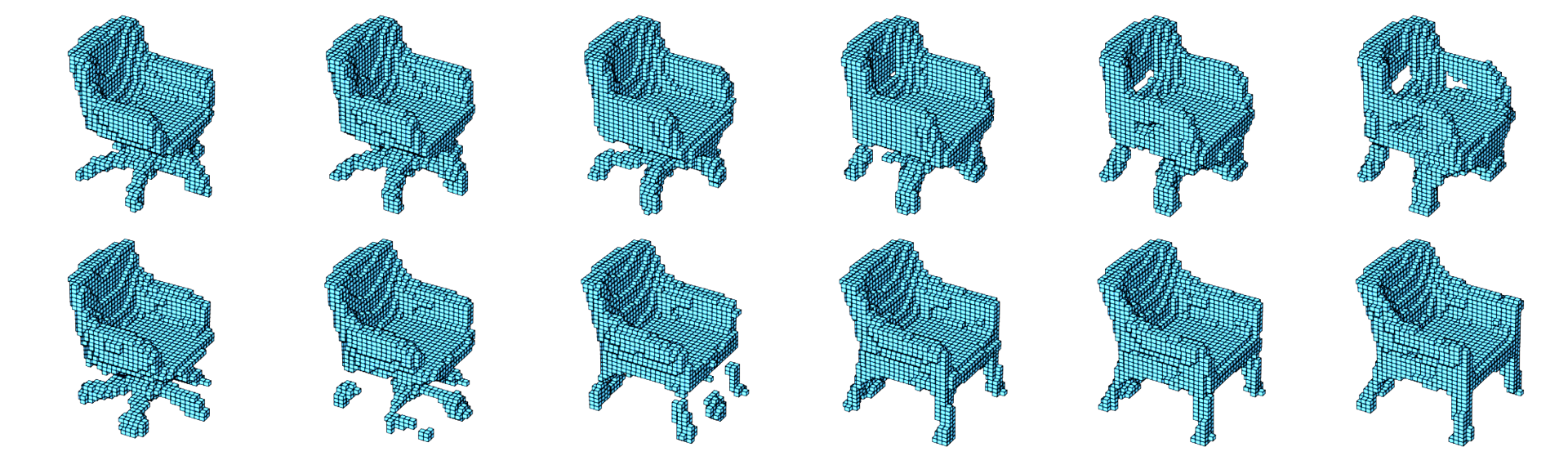}
        \caption{Top Row: Trained on SwitchVAE, changing the \enquote{chair leg type} feature also leads to the morphing of the top part. Bottom Row: Trained on Switch-BetaTCVAE with $\beta=5$, the top part stays more fixed while changing the \enquote{chair leg type} feature.}
        \label{fig:btcvae2}
    \end{subfigure}
    \caption{Disentangling latent features with Beta-TCVAE.}
    \label{fig:btcvae}
\end{figure}

\begin{figure*}[t]
    \centering
    \includegraphics[width=0.98\linewidth,trim=4 4 4 4,clip]{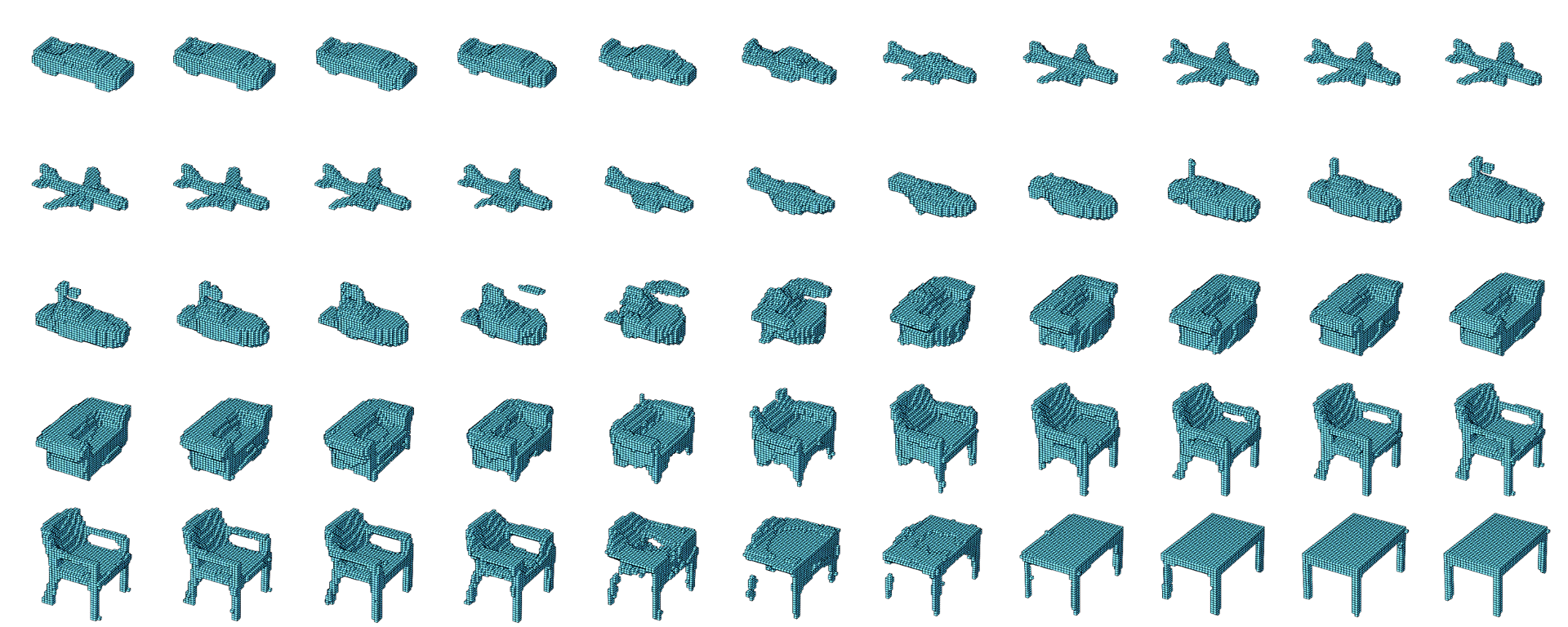}
    \caption{Shape interpolation between different categories.}
    \label{fig:interpolation}
\end{figure*}

\begin{figure}[t]
    \centering
    \begin{subfigure}[b]{\linewidth}
        \centering
        \includegraphics[width=0.8\linewidth,trim=4 4 4 4,clip]{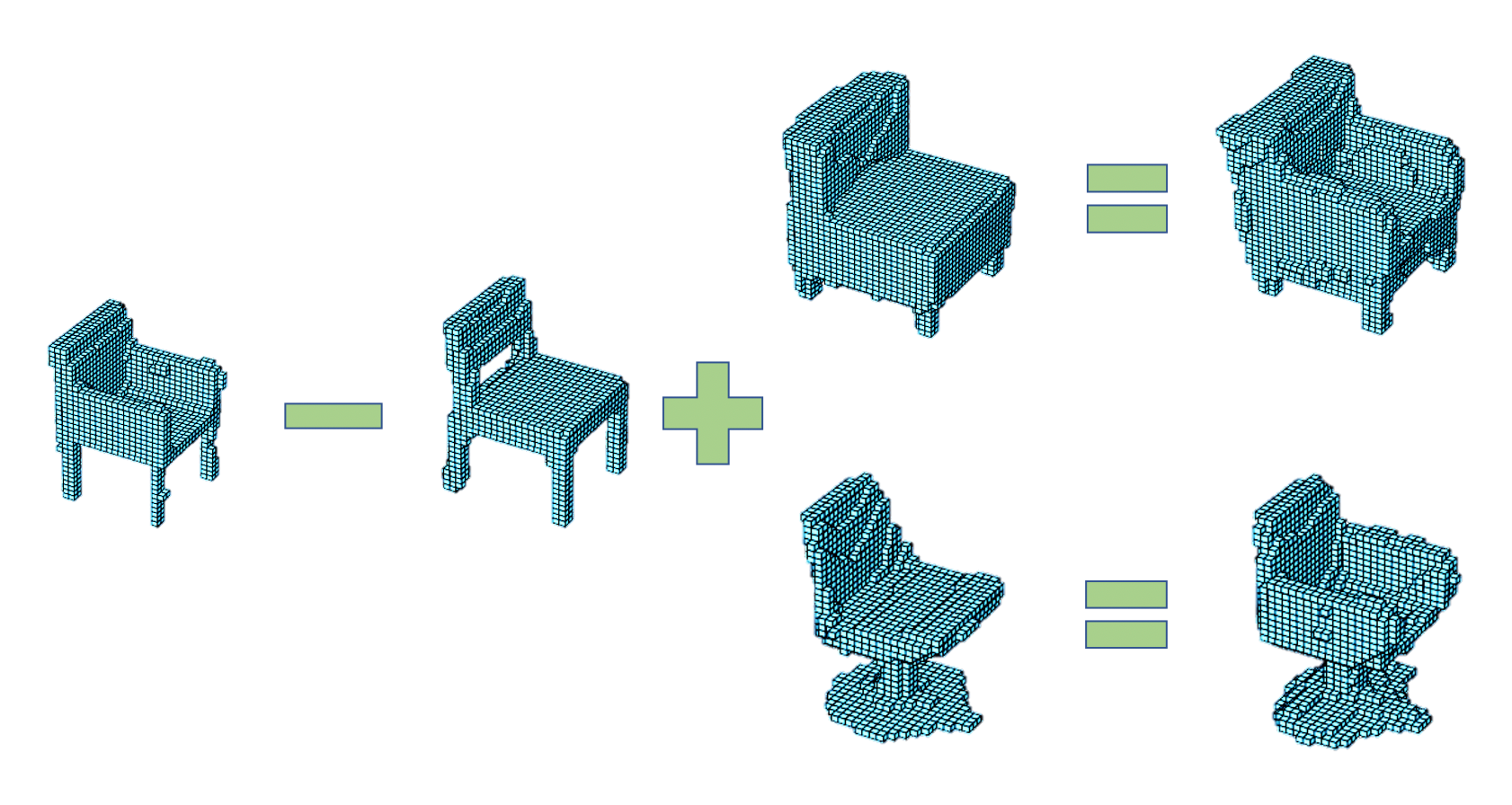}
        \caption{Shape arithmetic example with chair objects.}
        \label{fig:arithmetic_1}
    \end{subfigure}
    \begin{subfigure}[b]{\linewidth}
        \centering
        \includegraphics[width=0.8\linewidth,trim=4 4 4 4,clip]{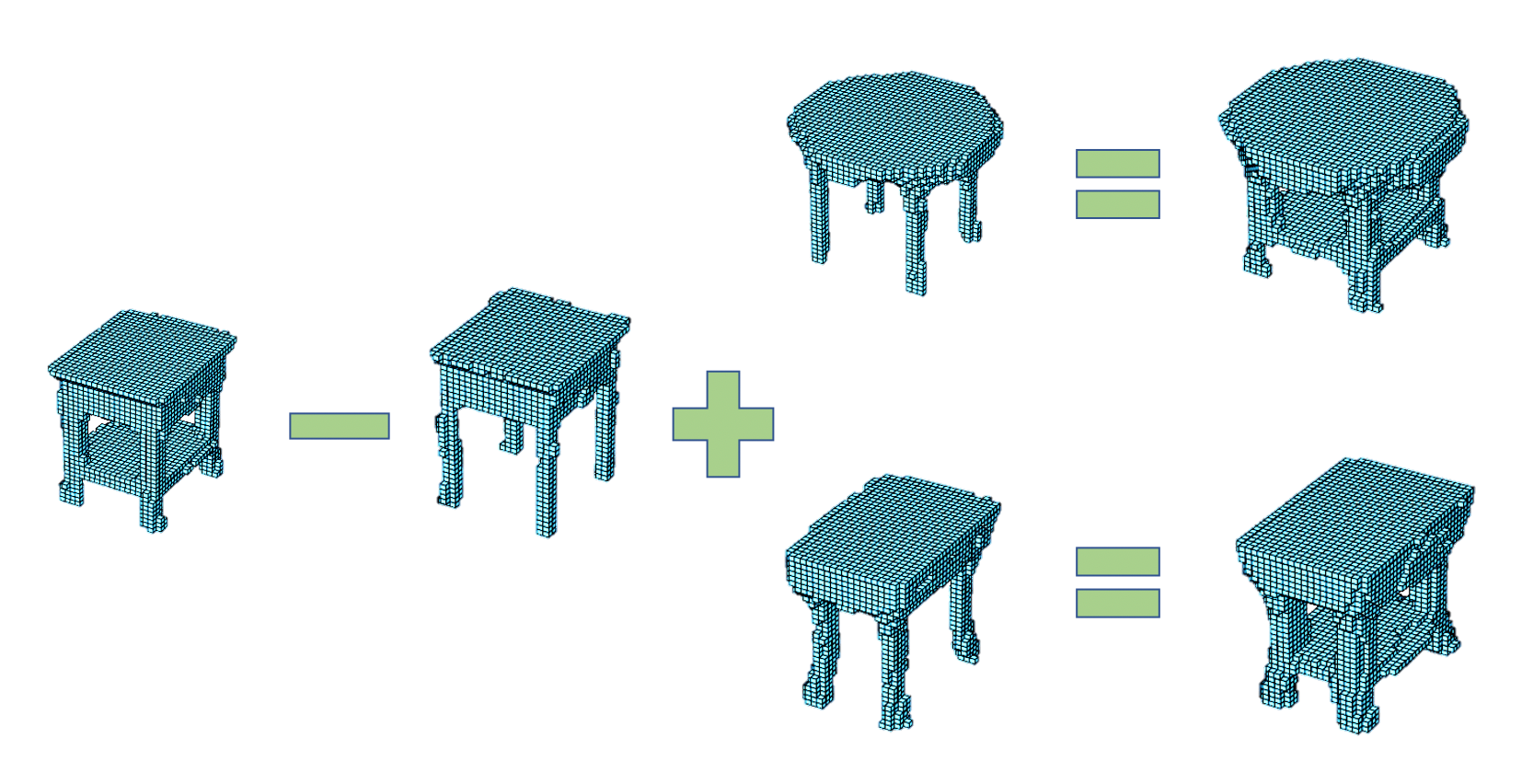}
        \caption{Shape arithmetic example with table objects.}
        \label{fig:arithmetic_2}
    \end{subfigure}
    \caption{Shape arithmetic for chairs and tables. (a) Adding chair arms to chair objects in the latent space. (b) Adding a middle layer to table objects in the latent space.}
    \label{fig:arithmetic}
\end{figure}

\subsection{Comparison to CrossPoint} 
\label{sec:crosspoint}
The closest work to ours is CrossPoint \cite{Afham2022CrossPointSC}, which uses images and point clouds as input for better point cloud latent representation learning. Apart from the different input data representations, our method differs from CrossPoint in three perspectives: (i) we use an additional reconstruction loss with the help of a decoder; (ii) we use VAE, other than AE, which introduces variance during the training; (iii) most importantly, we use a switching approach to enable the dynamic training of the framework. 

On the other hand, the encoder in CrossPoint is a plug-and-play module that can be easily replaced. We thus have replaced the point cloud encoder with a voxel encoder to conduct comparison experiments. The modified framework is illustrated in Figure \ref{fig:crosspoint}. Note that in the original framework of CrossPoint, it uses multi-point clouds and single-image as the input, while in our case, we use multi-images and single-volumetric data as the input. Hence for a fair comparison, the intra-modal contrastive loss is computed with image data in the modified framework. The numerical result is reported in Table \ref{table:3} under the method "CrossVoxel". We can observe that it achieves remarkable results, but not on par with ours. This is probably due to our framework (i) introduces another reconstruction loss with a decoder; (ii) allows latent representation variance using VAE; and (iii) uses more views of image during the training.

\subsection{Ablation Study on Switch Probability} 
\label{sec:switchProba}
One important parameter in our experiments is the switch probability, which decides the actual updating step ratio between two encoders. For a certain target data representation, e.g. the volumetric data, in order to improve the final performance of the voxel encoder for downstream tasks, the parameters of the voxel encoder should be updated more often. This means we should set the "switched probability" to the voxel encoder larger, compared to that of the image encoder. However, on the other hand, if the model focuses too much on the training of the voxel encoder and ignores or only trains slightly on the image encoder, it can hardly make use of the information from the image input. A trade-off between these two perspectives must be carefully made. Hence, an ablation study regarding the switch probability has been conducted and the numerical results are given in Table \ref{table:5}. We train a same SwitchVAE model but with different switch probabilities on the ModelNet40 dataset, and test it with the voxel data using the same classification benchmark in subsection \ref{sec:cls}. From it, we can see that a decent choice is setting the "switched probability" for the voxel encoder as $80\%$, while $20\%$ for the image encoder.

\subsection{Exploring Latent Representations}
This subsection showcases some qualitative results to give an indication that SwitchVAE training results in a superior latent representation that allows for better disentanglement between categories, as well as between the salient features of the 3D shapes in each category.

\textbf{Latent space interpolation:}
Similar to most 3D reconstruction papers, we also do the inter-class interpolation with our trained models as shown in Figure \ref{fig:interpolation}. It can be observed that our proposed method has the ability to perform a smooth transition between two shapes, even if they are from different categories.

\textbf{Shape arithmetic:}
Another way to explore the learned latent representations is to perform arithmetic operations in the latent space whilst observing their effect on the reconstructed geometry. We show some shape arithmetic results in Figure \ref{fig:arithmetic} with a model trained on the full 3D-R2N2 dataset. The model seems to capture the underline information and is capable of generating meaningful combined shapes that do not occur as 3D shapes in the original dataset.

\textbf{Feature disentanglement with VAE variations:}
One good thing with VAE models is that the latent space learned from it is more "meaningful" compared to that from GAN models. By tuning the value in one specific latent dimension, one can observe certain features on the output side changing smoothly. However, most features get entangled in multiple latent dimensions with the vanilla VAE. It has been reported that $\beta$-VAE \cite{Higgins2017betaVAELB} and $\beta$-TCVAE \cite{Chen2018IsolatingSO} can produce better disentangled features in the latent space. We merge it with our proposed method into SwitchBTCVAE. We train our model with $\beta=5$ on the chair category with a same number of epochs as the other experiments. Although a small decrease in the reconstruction performance metrics is observed, by investigating the learned latent representations, we find that some features have been better disentangled as shown in Figure \ref{fig:btcvae}.

\section{Conclusion and Outlook}
\label{sec:conclusion}
In this paper, we propose a generative-contrastive learning pipeline for learning better latent representations for 3D volumetric shapes, with the help of additional modality input. The switching approach makes the joint training for both encoders possible with competitive reconstruction results. Classification experiments on ModelNet have also been carried out to validate the effectiveness of the proposed method. Improved classification results indicate that better latent representations have been learned with our proposed SwitchVAE architecture.

For future directions, other 3D data modalities, e.g., point clouds and meshes may also be used. A new contrastive loss may be designed and an optimal switching policy may be studied. More research may be conducted to make the latent presentations more feature-disentangled or more interpretable. In our experiments, although an additional contrastive loss has been applied, we still observe large distances between latent representations generated from the different input formats. A thorough study of how to force a closer distance between representations from different formats without leading to a collapse due to increased contrastive loss could provide deeper insights into the fundamental principles of contrastive learning.

{\small
\bibliographystyle{ieee_fullname}
\bibliography{main}
}

\begin{IEEEbiography}[{\includegraphics[width=1in,height=1.25in,clip,keepaspectratio]{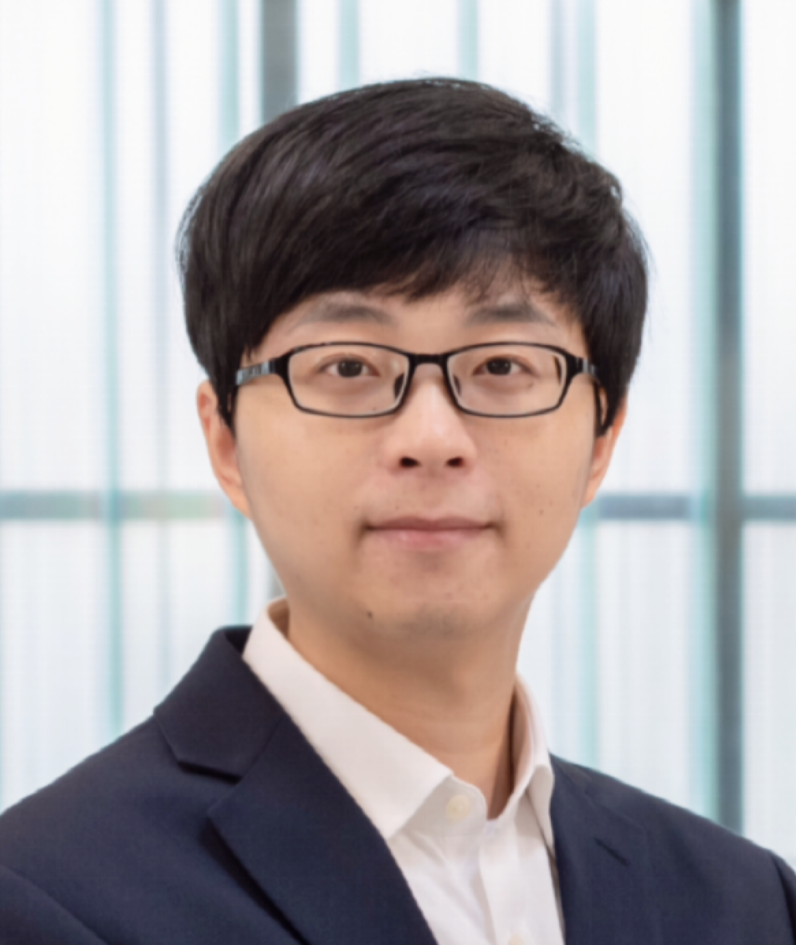}}]{Chengzhi Wu}
Chengzhi Wu received both B.E. and M.S. degrees from Nanjing University, Nanjing, China. He is currently pursuing a Ph.D. degree in the Department of Computer Science at Karlsruhe Institute of Technology (KIT), Karlsruhe, Germany. His research interests include machine learning and optimization algorithms, robust control of dynamic systems,  deep learning, computer vision for remanufacturing, and 3d data analysis.
\end{IEEEbiography}

\begin{IEEEbiography}[{\includegraphics[width=1in,height=1.25in,clip,keepaspectratio]{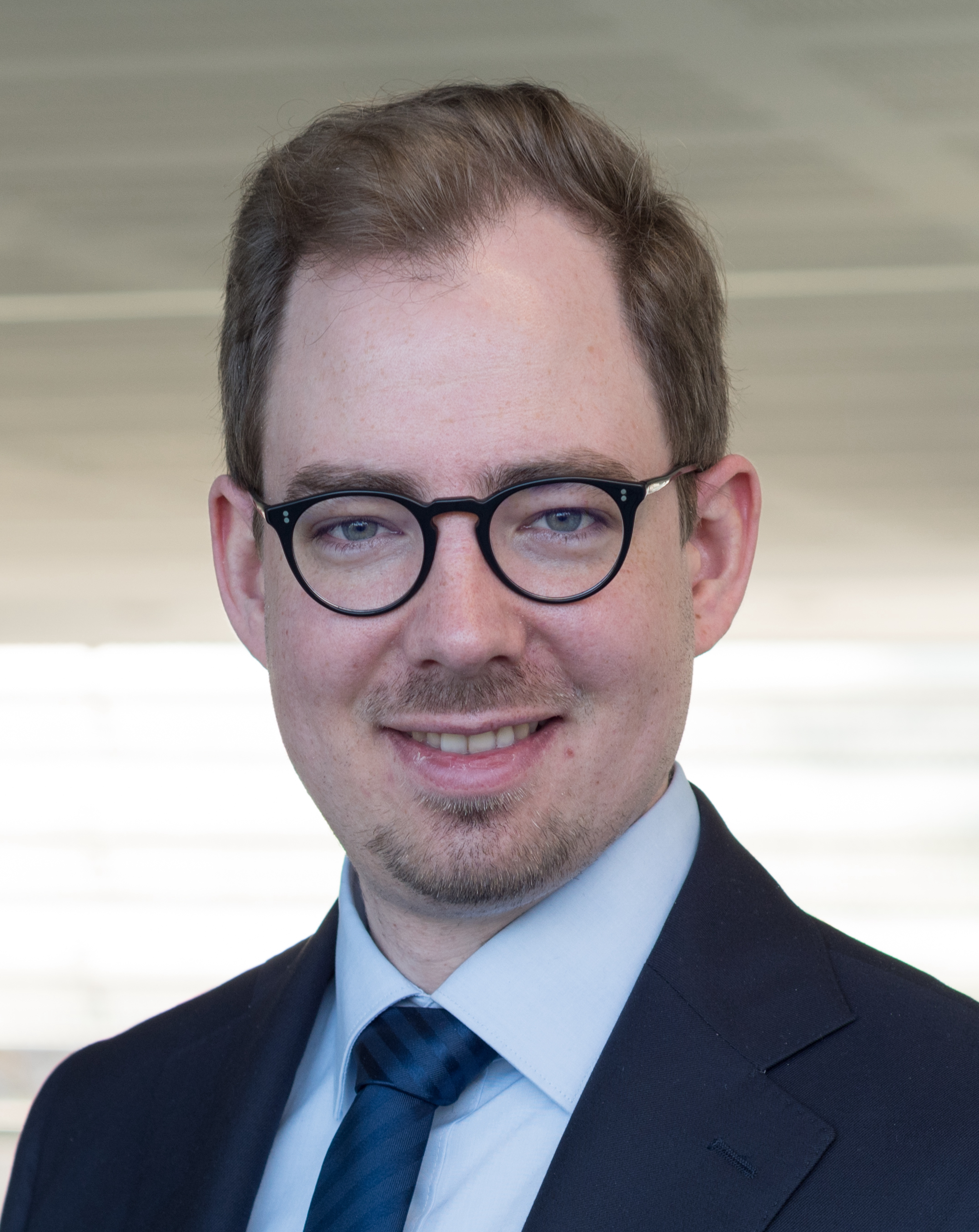}}]{Julius Pfrommer}
Dr.-Ing. Julius Pfrommer leads the Department for Cognitive Industrial Systems at Fraunhofer IOSB. He holds engineering degrees from Karlsruhe Institute of Technology (KIT) and the Institut National Polytechnique in Grenoble. His PhD in computer science (summa cum laude) was awarded at KIT. His research interests include distributed systems, planning under uncertainty, and the use of machine learning for modeling, optimization and control in cyber-physical systems.
\end{IEEEbiography}

\begin{IEEEbiography}[{\includegraphics[width=1in,height=1.25in,clip,keepaspectratio]{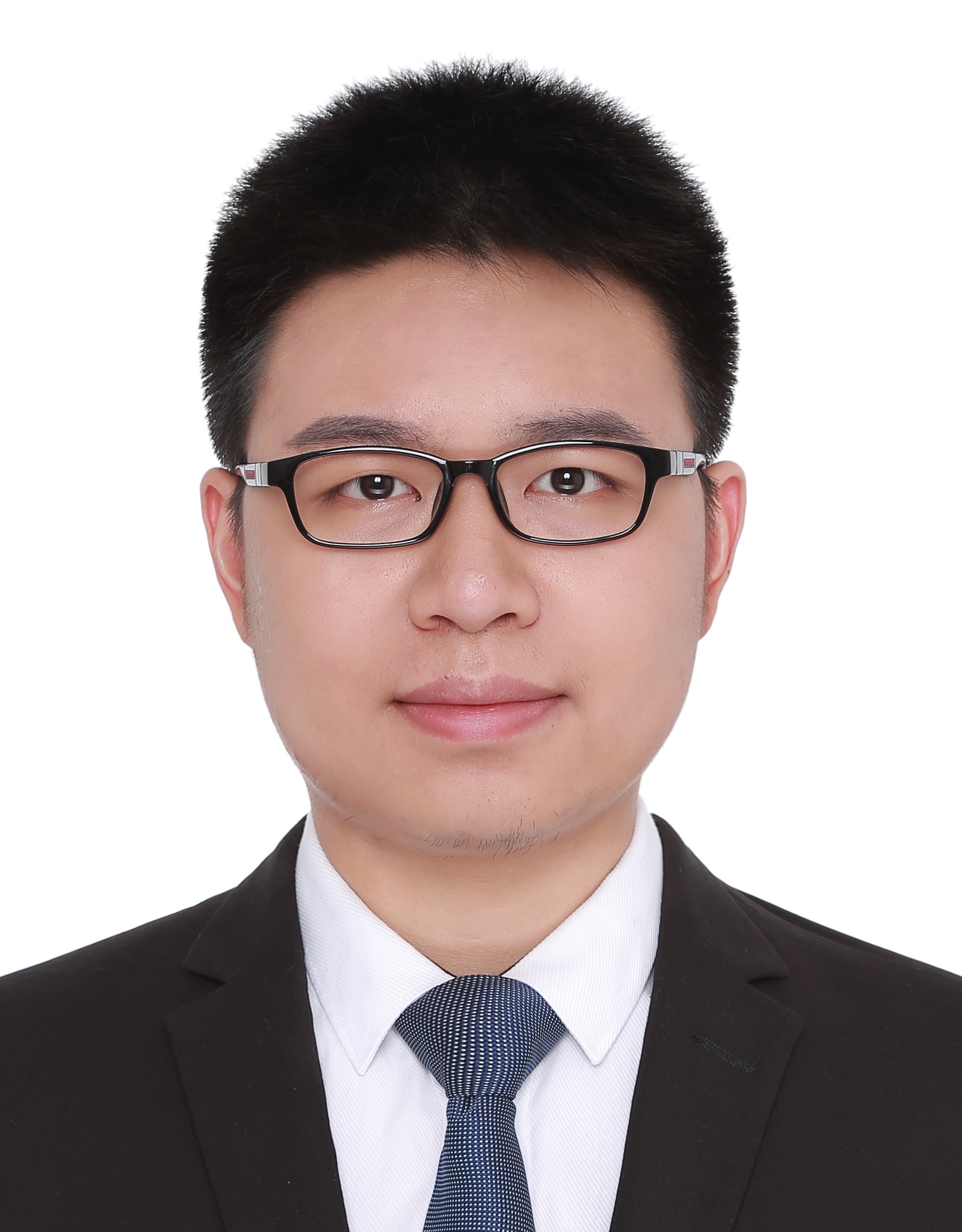}}]{Mingyuan Zhou}
Mingyuan Zhou received the B.E. degree from the Beijing Jiaotong University, Beijing, China, and the M.S. degree from the Karlsruhe Institute of Technology (KIT), Karlsruhe, Germany. He is currently pursuing the Ph.D. degree with the Internet of Things thrust in The Hong Kong University of Science and Technology (Guangzhou), China. His research interests include deep learning, mechanical engineering, and structural engineering.
\end{IEEEbiography}

\begin{IEEEbiography}[{\includegraphics[width=1in,height=1.25in,clip,keepaspectratio]{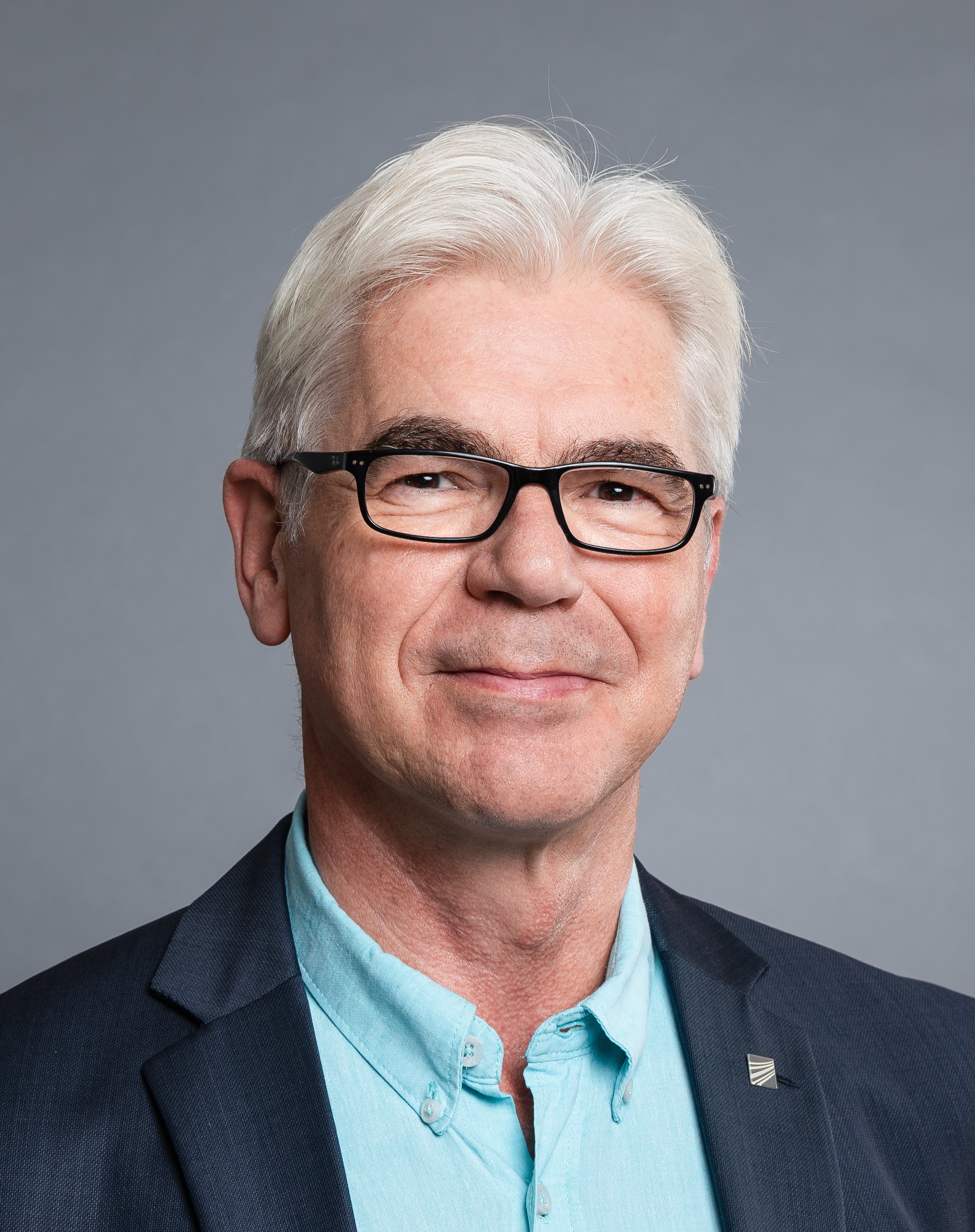}}]{Jürgen Beyerer}
Prof. Dr.-Ing. Jürgen Beyerer has been a full professor for informatics at the Institute for Anthropomatics and Robotics at the Karlsruhe Institute of Technology KIT since March 2004 and director of the Fraunhofer Institute of Optronics, System Technologies and Image Exploitation IOSB in Ettlingen, Karlsruhe, Ilmenau, Görlitz, Lemgo, Oberkochen and Rostock. Research interests include automated visual inspection, signal and image processing, variable image acquisition and processing, active vision, metrology, information theory, fusion of data and information from heterogeneous sources, system theory, autonomous systems and automation.
\end{IEEEbiography}

\vfill

\end{document}